# A Context Alignment Pre-processor for Enhancing the Coherence of Human–LLM Dialog

DING WEI

College of Architecture, Nanjing Tech University

kxrdk@163.com

**Abstract:** Large language models (LLMs) have made remarkable progress in generating fluent text, but they still face a critical challenge of contextual misalignment in long-term and dynamic dialogue. When human users omit premises, simplify references, or shift context abruptly during interactions with LLMs, the models may fail to capture their actual intentions, producing mechanical or off-topic responses that weaken the collaborative potential of dialogue. To address this problem, this paper proposes a computational framework called the Context Alignment Pre-processor (C.A.P.). Rather than operating during generation, C.A.P. functions as a pre-processing module between user input and response generation. The framework includes three core processes: (1) semantic expansion, which extends a user instruction to a broader semantic span including its premises, literal meaning, and implications; (2) time-weighted context retrieval, which prioritizes recent dialogue history through a temporal decay function approximating human conversational focus; and (3) alignment verification and decision branching, which evaluates whether the dialogue remains on track by measuring the semantic similarity between the current prompt and the weighted historical context. When a significant deviation is detected, C.A.P. initiates a structured clarification protocol to help users and the system recalibrate the conversation. This study presents the architecture and theoretical basis of C.A.P., drawing on cognitive science and Common Ground theory in human-computer interaction. We argue that C.A.P. is not only a technical refinement but also a step toward shifting human-computer dialogue from one-way command-execution patterns to two-way, self-correcting, partnership-based collaboration. Finally, we discuss implementation paths, evaluation methods, and implications for the future design of interactive intelligent systems.

## 1 INTRODUCTION

Since the advent of the transformer architecture, such large language models (LLMs) as the GPT series and LlaMA have emerged as the most transformative force in natural language processing [1, 2]. They have demonstrated remarkable capabilities in tasks such as the generation, summarization, and translation of text as well as answering questions, and are being used increasingly commonly as tools for human knowledge workers and creators. However, as human–machine interactions evolve from simple, single-turn questions and answers (Q&A) to complex, long-term, and multi-turn collaborative dialogs, a deep limitation of LLMs has gradually emerged: contextual misalignment.

Human conversation is inherently an efficient but "uncertainty-rich" collaborative process. Participants rely on shared knowledge, common focal points, and a continually updated "common ground" [3] to comprehend each other's implied premises, simplified references, and non-linear leaps in reasoning. However, currently available LLMs remain largely "faithful but naïve" executors. They primarily rely on limited contextual windows and attention mechanisms to interpret user input. When users issue seemingly simple commands that imply complex historical contexts or subtle shifts in intent, the models often deviate due to their inability to dynamically track deeper contextual nuances. This deviation manifests in the following forms:

(1) Mechanical responses: Models strictly execute instructions based on their literal meaning, while ignoring their true purpose within the given conversational flow.

(2) Focus drift: The model is completely derailed from the established conversational thread by new instructions, which results in logical disconnects.
(3) Missed opportunities: The model fails to recognize the creative potential of or invitations for deeper exploration that are embedded in new user instructions, and maintains the conversation at a superficial level.

The above issues not only reduce the efficiency of interaction, but, more critically, also hinder the formation of genuine "intellectual partnerships" between humans and LLMs. To address this core challenge, the relevant research has primarily focused on expanding context windows, optimizing attention mechanisms, and enhancing model compliance through instruction tuning [4]. However, these approaches are largely "passive" adaptations that fail to fundamentally endow models with the ability to actively calibrate the context.

This paper proposes a solution to the above problem called the Context Alignment Pre-processor (C.A.P.). The C.A.P. is a lightweight and modular computational framework that operates prior to the LLM's primary task (response generation), with the sole objective of ensuring that the model and user are "on the same page" at the given time step. By simulating the human conversational mechanisms of reflection and confirmation, it dynamically evaluates the consistency between the given instructions and the history of the dialog. Once it detects a potential "misalignment," it pauses generation and initiates a clarification protocol, thus transforming the burden of "guessing" into an explicit and collaborative contextual calibration with the user.

The main contributions of this paper are as follows:

(1) It introduces a computational framework (C.A.P.) that is specifically designed to proactively manage and calibrate the understanding of the conversational context by the LLM before response generation.
(2) It formally defines the three core components of the C.A.P.—semantic expansion, time-weighted context retracing, and verification of alignment with decision branching—and elucidates their collaborative mechanism.
(3) It grounds the C.A.P. framework in robust theoretical foundations from cognitive science and Human-Computer Interaction (HCI), particularly its connection to the Common Ground theory and mechanisms of conversational repair. This establishes its theoretical legitimacy.
(4) This paper explores pathways for the implementation of the C.A.P, metrics for its evaluation, and directions for future research in the area. This offers a feasible technical roadmap to elevate LLMs from "tools" to "partners."

The remainder of this paper is structured as follows: Section 2 reviews related work in the field, Section 3 details the architecture and workflow of the C.A.P. framework, while Section 4 explores its theoretical foundations and deeper implications. Section 5 describes methods for implementing and evaluating the C.A.P, Section 6 discusses its potential impacts and limitations, and Section 7 summarizes the conclusions of this paper.

## 2 RELATED WORK

The motivation for research on the C.A.P. framework and its design philosophy are closely related to current research in three domains: context processing for LLMs, dialog management systems, and foundational theories of human–computer interaction.

### 2.1 Context Handling and Limitations of LLMs

Modern LLMs are built upon the transformer architecture, the self-attention mechanism of which enables them to weigh the importance of parts within the input sequences [1]. Theoretically, this allows the models to capture long-range dependencies. In practice, however, the contextual understanding of LLMs remains constrained by several factors. The first is the finite context window. Although window sizes are increasing (from thousands to millions of tokens), they ultimately face physical limits. Critical early information for extremely long dialogs may thus be discarded. The second constraint on the window size is the "lost in the middle" phenomenon. Research suggests that when processing long inputs, LLMs focus primarily on information at the beginning and end, such that the middle sections can be easily overlooked [5]. Finally, the recency bias causes models to overemphasize recent rounds of dialog while potentially overlooking early groundwork that sets the tone for the overall conversation. The C.A.P.'s time-weighted backtracking mechanism specifically counters this bias by algorithmically forcing the model to revisit and evaluate the importance of the historical context.

### 2.2 Dialog Management

The Dialog Manager handles Dialog State Tracking (DST) in traditional task-oriented dialog systems [6]. DST aims to accurately represent the user's intent and the slot values in each time step based on the dialog history. However, traditional DST primarily applies to well-defined and domain-restricted tasks (e.g., booking tickets and weather-related queries). For open-domain and creative collaborative dialogs, the user's "intent" is fluid and emergent, and is difficult to characterize by using predefined slots. Recent research has attempted to leverage the LLMs themselves for dialog state management [7], but this approach often couples it with the generation task, such that a dedicated mechanism for "reflective" context alignment is lacking. The C.A.P. can be viewed as a novel and lightweight dialog manager that tracks higher-level "semantic coherence," rather than specific "slots," and can proactively interrupt and repair when incoherence is detected.

### 2.3 Common Ground in Human–Computer Interaction (HCI)

The "common ground" theory, proposed by Clark and Brennan, pertains to the knowledge, beliefs, and assumptions shared by the participants of a dialog [3]. Establishing and maintaining common ground is crucial for successful communication. When one party perceives potential deviations in the common ground, they initiate "repair mechanisms," such as requesting clarification ("Did you mean ...?"). In human–computer interaction, enabling machines to effectively participate in the construction of a common ground remains a core challenge [8]. Current research has primarily focused on enabling systems to generate more "context-aware" responses, such as by referencing prior dialogic content. However, these approaches are passive. The C.A.P.'s uniqueness lies in its explicit algorithmic formalization of repair mechanisms. Its "alignment check" process simulates the computational assessment of the stability of a shared ground, while its "clarification protocol" directly borrows from human conversational repair mechanisms. This endows the AI with the unprecedented capability of acknowledging that it may have "lost track" and requesting assistance from its human partner. This marks a significant shift from pursuing "omniscient" AI toward pursuing "honest, collaborative" AI.

In summary, the C.A.P. framework fills a critical gap in prevalent research by introducing a pre-processing stage that is independent of the generation task, and by explicitly simulating human conversational reflection and repair mechanisms. This provides LLMs with a structured method for actively managing and calibrating the conversational context.

## 3 DETAILED EXPLANATION OF C.A.P. FRAMEWORK

The core design philosophy of the C.A.P. is to "think before acting." Upon receiving a new user instruction $A$ at time point $T_A$, it does not immediately pass it to the LLM for generation. Instead, it initiates a preprocessing task comprising three sequential processes.

### 3.1 Overall Framework Architecture

The C.A.P. functions as middleware, and is positioned between the user and the LLM's core generation module. Its workflow is as follows:

    (1) Input: Real-time request $A$ submitted by the user at time point $T_A$.
    (2) C.A.P. Processing:
        Process 1: Semantic expansion.
        Process 2: Time-weighted context retrieval.
        Process 3: Alignment check and decision branching.
    (3) Output:
        If aligned: Pass the original instruction $A$ (possibly with a context summary appended) to the LLM generation module.
        If misaligned: Suspend the main task and present the user with the "Clarification Protocol" interface.

### 3.2 Process One: Semantic Expansion

This process is designed to overcome the limitation of literal interpretation of user instructions by the LLM. It expands a single instruction $A$ into a set $Set(A)$ that encompasses its potential semantics, thereby transforming a "point-like" instruction into an "interval-like" semantic space.

$$Set(A) = \{A^-, A, A^+\}$$

**$A$ (Literal):** This is the user's original instruction, serving as the center point of the semantic space.

**$A−$ (Prerequisite/Foundation)**: This constitutes implicit prerequisites, foundational definitions, or a more specific version required to execute command $A$. For example, if $A$ is "Provide the formula for the functional synergy index of each district in a city," $A−$ may be "First define what constitutes functional complementarity and activity correlation among districts in a city."

**A+ (Implication/Application)**: This is the logical extension, scenario of application, or a broader and more exploratory version of instruction $A$. For example, if $A$ is "Provide a formula for the functional synergy index of urban districts," A+ could be "Explore how this index can be used to construct urban functional networks and perform a community analysis of urban districts."

    $A−$ and A+ can be generated through a single, small LLM invocation by using such meta-prompts as "What prerequisites are needed to execute this instruction?" and "What is the next step for this instruction?" This step aims to capture broader semantic associations for the subsequent verification of alignment.

### 3.3 Process Two: Time-weighted Context Retrieval

This process simulates the focal nature of human memory, in which recent dialogic content is typically most relevant. However, the crucial early context should not be forgotten either. It retrieves a weighted context subset from the complete dialog history.

$$H = \{H1, H2, ..., Hn\}$$

(1) **Retrieval**: Extract the *k* most recent rounds of dialogic history, e.g.,

$$\text{Hcontext} = \{A1, R1, ..., Ak, Rk\}$$

where *Ai* represents user instructions and *Ri* denotes model responses.

(2) **Weighting**: Assign a weight *Wi* to each historical dialogic round *Hi* (or each instruction). This weight is a decreasing function of its temporal distance from the current time *TA*. A simple yet effective function to this end is the inverse proportional function:

$$Wi = f(TA - Ti) = (TA - Ti)/T + 1$$

where *Ti* is the timestamp of the historical instruction *Hi*, and $\tau$ is a temporal scale-related parameter that controls the rate of decay of the weights. This formula ensures that the largest weight is assigned to the most recent dialogic turn, while earlier turns exhibit a smooth, non-zero weight decay.

### 3.4 Process Three: Alignment Check and Decision Branching

The C.A.P. is responsible for making the final decision. It does so by calculating the alignment score between the semantic space *Set(A)* of the current prompt and the weighted historical context.

**(1) Vectorization**: By using a pre-trained model of sentence embedding (e.g., Sentence-BERT [9]), convert each element in *Set(A)* (A−, A, A+), and each instruction *Hi* in the weighted history into a high-dimensional semantic vector *v(x)*.

**(2) Calculate Alignment Score**: The alignment score *Salign* is defined as the maximum weighted similarity between *Set(A)* and the historical context:

$$S_{\text{align}}(A, H) = \max_{a \in \text{Set}(A)} \left( \sum_{i=1}^{k} W_i \cdot \text{sim}(v(a), v(H_i)) \right)$$

where sim(*v1*, *v2*) is the cosine similarity. It measures the extent to which the current instruction (and its latent semantics) can be "explained" by the most recent and relevant dialog history.

**(3) Decision Branch**: Compare the computed alignment score *Salign* with a preset threshold $\theta$.

**If *Salign* ≥ $\theta$ (Alignment Confirmed):**

**Conclusion**: The current instruction *A* is a natural continuation of the flow of the conversational logic.
**Action**: Execute normally. Pass instruction *A* to the LLM core generation module. To further enhance coherence, inject the most similar historical entry $H_j$ into the prompt as additional context.

**If *Salign* < $\theta$ (Misalignment Alert):**

**Conclusion**: A potential jump in context or ambiguous instruction has been detected. This means that the user intent may have shifted significantly, or a simplified expression exceeds the boundaries of safe inference.
**Action**: Initiate the Clarification Protocol. Pause the primary task and present the user with a structured interface:
**Repeat**: "Your current real-time request is: '[Repeat instruction A].'"
**Alert**: "I note that this request appears substantially different in subject matter from our previous discussion of '[Repeat most similar historical instruction *Hj*].'"

**Empower**: "To better understand your intent, I need your assistance. Would you like to:"

**Offer Choices**: a) Proceed with this new request; b) Correct my understanding—your request is actually a deepening or variation of the previous topic; c) Alternatively, provide a clearer new request.

This protocol is designed to politely and non-confrontationally return control to the user and collaboratively restore a "shared foundation."

## 4 THEORETICAL FOUNDATIONS AND SIGNIFICANCE

The C.A.P. framework is not merely an engineering solution; it is deeply rooted in the theoretical foundations of cognitive science and human–computer interaction, which endows it with significance beyond pure technical optimization.

### 4.1 From Cognitive Science: Simulating Human Reflection and Repair

Human dialog is far from a perfect linear process. It is filled with interruptions, corrections, and clarifications. These "disruptions" are precisely the key mechanisms ensuring successful communication. When one party in a conversation is uncertain about their understanding of the meaning of the other, they instinctively pause and seek confirmation through questioning, paraphrasing, or other means. This is a form of metacognitive ability as the awareness of one's own cognitive state.

The C.A.P.'s "alignment check" is a computational simulation of this **metacognitive** reflection. It prevents AI from being overconfident, and teaches it to practice "self-doubt." The "clarification protocol" directly implements **dialog** repair mechanisms. Through this process, AI transforms from a passive information processor into an active participant in communication. It can identify potential barriers to communication and invite its human partner to collaboratively overcome them.

### 4.2 From an HCI Perspective: Building and Maintaining a "Shared Ground"

As previously noted, shared ground is the cornerstone of collaborative activities. Clark and Brennan [3] have noted that different communication media carry varying "grounding costs" in supporting the construction of shared ground. Face-to-face human interaction incurs the lowest cost, as the participants can rapidly confirm a common understanding through multiple channels, like eye contact and gestures. By contrast, text-based human–computer interaction entails significantly higher costs for establishing a shared ground.

The C.A.P. framework can be viewed as a mechanism designed to reduce the costs of grounding in human–computer dialog. When it detects potential instability in grounding (i.e., low alignment scores), it rapidly rebuilds consensus through a low-cost clarificatory interaction, thereby avoiding the substantial sunk costs of subsequent rounds of dialog caused by misunderstanding. From this perspective, the C.A.P. carves out an efficient path for maintaining a shared foundation between humans and machines within the limited text-based channel of interaction.

### 4.3 Paradigm Shift: From "Tool" to "Partner"

The ultimate significance of the C.A.P. lies in the fact that it represents a paradigm shift in human–machine relations.

**Tool Paradigm**: AI acts as a passive executor, and humans bear the responsibility of issuing clear and unambiguous instructions. The burden of communication thus rests entirely on the human side.

**Partner Paradigm**: AI acts as an active collaborator. It recognizes ambiguities in communication and shares the responsibility for clarification with humans. Communication then becomes bidirectional, and is jointly constructed.

By endowing AI with the capabilities of "reflection" and "seeking assistance," the C.A.P. enables patterns of AI behavior that closely resemble those of a true conversational partner. This partnership is built on trust, which stems from AI's ability to acknowledge its limitations and commit to achieving a deep understanding in its collaboration with humans.

## 5 PATHWAYS OF IMPLEMENTATION AND EVALUATION

As a conceptual framework, the value of the C.A.P. ultimately requires demonstration through its implementation and rigorous evaluation.

### 5.1 Path of Implementation

The C.A.P. can be implemented as a standalone Python library or an API service, so that it can encapsulate calls to the underlying LLMs (e.g., the GPT-4 API).

- **(1) Storage of Conversation History**: A simple in-memory queue can be used to this end. Vector databases (e.g., Pinecone, Chroma) can store embeddings of historical conversations for efficient retrieval for applications that require persistence.
- **(2) Semantic Expansion**: This is achieved by sending carefully crafted metaprompts to the same LLM or another, smaller LLM.
- **(3) Vectorization**: Efficient models of sentence embedding, like all-MiniLM-L6-v2, are recommended for this task owing to their balanced performance and speed.
- **(4) Parameter Tuning**: Key parameters of the framework, like the coefficient of time decay $\tau$ and threshold of alignment $\theta$, require tuning through experiments on benchmark datasets. Setting the value of $\theta$ is particularly critical because too high a value can cause excessive clarification such that this impairs the fluency of the LLM, while too low a value reduces the effectiveness of its "alert" function.

### 5.2 Methods of Evaluation

Evaluating the C.A.P.'s effectiveness requires a multi-dimensional framework that combines quantitative and qualitative metrics.

**A/B Testing**: This serves as the core method of evaluation. Recruit a group of users to complete a series of complex, multi-round collaborative tasks (e.g., jointly developing a business plan, writing a short story) by using two versions of the system. Control group: Users interact directly with the base LLM. Experimental group: Users interact with the LLM integrated with the C.A.P.

**Quantitative metrics:**

**Task Success Rate**: It measures the extent to which the user completes predefined tasks.
**Dialogic Efficiency**: It is the total number of rounds or total time required to complete the task. We anticipate that the C.A.P. may require more rounds (due to clarifications) but can reduce the total time wasted due to misunderstandings.
**Frequency of Clarification**: It is the number of times that the C.A.P. triggers the clarification protocol.

**User Satisfaction**: It is assessed by using standardized questionnaires, such as the System Usability Scale (SUS) [10] or the PARADISE framework [11], to evaluate the users' subjective perceptions of the quality of interaction, and the intelligence and cooperativeness of the system.

**Qualitative Metrics:**

**Conversation Analysis**: It involves qualitatively coding transcribed dialogic texts to analyze the occurrence of "catastrophic misunderstandings"—instances where users express frustration—and moments reflecting "deep collaboration."

**Post-task Interview**: It involves conducting semi-structured interviews with the users to gain insights into their perception of differences between the systems in terms of "understanding," "sense of cooperation," and "trust."

We anticipate that a system that incorporates the C.A.P. will significantly outperform the baseline system on key metrics, including the rate of task success, user satisfaction, and "sense of collaboration."

## 6 DISCUSSION AND LIMITATIONS

The C.A.P. framework offers a promising path for enhancing the quality of human–computer dialog, but its implementation and application remain limited, and face several challenges.

First, the computational overhead of the framework requires consideration. Each preprocessing step in the C.A.P., particularly the invocation of the LLM for semantic expansion and vector computations, increases its response latency. Optimizing the C.A.P.'s efficiency of execution without significantly impacting its fluency of interaction remains a critical engineering challenge.

Second, parametric sensitivity—especially the threshold of alignment $\theta$—is critical to user experience. A fixed threshold may fail to accommodate all users and types of dialogs. Future research should explore dynamic mechanisms of threshold adjustment, such as automatically adapting $\theta$ based on the domain of the dialog, user expertise, or historical patterns of interaction.

Third, the design of the clarification protocol requires refinement. Excessively frequent or poorly designed clarifications may annoy users with "over-interruption." Designing clarification-related interactions that are both effective and natural is an HCI problem that requires iterative optimization through extensive user research.

Finally, the C.A.P. primarily addresses semantic coherence, and has a limited capability for deeper "alignment" involving emotions, values, or complex social dynamics. While it represents an important starting point for such investigation, it is far from the endpoint of complete human–AI alignment. Despite these limitations, we think that the design philosophy embodied by the C.A.P., which empowers AI with self-reflection and the initiative to seek assistance, holds profound value. It encourages us to rethink the essence of intelligence, and move beyond the pursuit of raw performance to prioritize the authenticity of human–AI collaboration.

## 7 CONCLUSION

In an era of deepening human–LLM integration, the quality of human–machine dialog directly determines the upper limit of collaborative creation. This paper has considered the pervasive issue of "context misalignment" in long-term conversations involving LLMs, and has proposed a computational framework to solve the problem called the "Context Alignment Preprocessor" (C.A.P.).

By introducing three core processes—semantic expansion, time-weighted context recall, and alignment verification—

prior to text generation, the C.A.P. endows AI with the unprecedented capability to actively assess its own understanding of user intent and, upon detecting potential misalignment, request its human partner to jointly calibrate the interaction. This framework represents not merely a technical optimization, but a profound paradigm shift that enables human–computer interaction to evolve from unidirectional command execution toward a bidirectional, self-repairing collaborative partnership.

By grounding C.A.P. in robust theories of cognitive science and human–computer interaction, and clearly outlining pathways for its implementation and evaluation, C.A.P. provides a sound foundation for building smarter, more reliable, and more trustworthy next-generation interactive AI systems. Future work in the area should focus on the engineering implementation of this framework, large-scale user studies, and continual optimization of its core algorithms. The ultimate goal is to ensure that every conversation between humans and AI becomes a truly meaningful resonance of ideas.